\documentclass[twoside]{article}
\usepackage{aistats2014}
\usepackage{multirow}
\usepackage{multicol}
\usepackage{url}
\usepackage{lipsum}
\usepackage[algo2e,ruled]{algorithm2e}
\usepackage{graphics,verbatim,array}
\usepackage{algorithm,algorithmic}
\usepackage{amsmath}
\newcommand{\eat}[1]{}
\DeclareMathOperator*{\argmax}{arg\,max}

\begin{document}

\input{amssymb.mac}

\def\bfy{{\bf y}}
\def\bfby{{\bf {\bar y}}}
\def\bfyc{{{\bf y}_c}}
\def\bfbyci{{{\bfby}^{(i)}_c}}
\def\bfyci{{{\bfy}^{(i)}_c}}
\def\brbfyc{{{\bf {\bar y}}_c}}
\def\barc{{\bar c}}
\def\bfybarc{{{\bf y}_{\bar c}}}
\def\bfyccbar{{{\bfy_{c,{\barc}}}}}
\def\bfbaryj{{{\bfy}_{c,-j}}}
\def\bfx{{\bf x}}
\def\bfym{{{\bf y}_m}}
\def\bfyms{{\bf y}^*_m}
\def\bfyo{{{\bf y}_o}}
\def\bfxi{{\bf x}^{(i)}}
\def\bfyi{{{\bf y}^{(i)}}}
\def\bfymi{{{\bf y}^{(i)}_m}}
\def\bfyoi{{{\bf y}^{(i)}_o}}
\def\bfyoj{{{\bf y}^{(j)}_o}}
\def\bfymj{{{\bf y}^{(j)}_m}}
\def\bfyci{{{\bf y}^{(i)}_c}}
\def\bfbyci{{{\bf {\bar y}}^{(i)}_c}}
\def\bfyca{{{\bf y}_{c_a}}}
\def\bfybca{{{\bf y}_{{\bar c}_a}}}
\def\bfycb{{{\bf y}_{c_b}}}
\def\bbfyca{{{\bar {\bf y}}_{c_a}}}
\def\bbfycb{{{\bar {\bf y}}_{c_b}}}
\def\bfybc{{{\bf y}_{\bar c}}}
\def\bfyMbc{{{\bf y}_{-\barc}}}
\def\si{{s(\bfyi)}}
\def\sci{{s_c({\bfyci};\theta_c)}}
\def\scbi{{s_c({\bfbyci};\theta_c)}}
\def\s{{s(\bfy)}}
\def\sym{{s(\bfym;\theta_m)}}
\def\syo{{s(\bfyo;\theta_o)}}
\def\symi{{s(\bfymi;\theta_m)}}
\def\syoi{{s(\bfyoi;\theta_o)}}
\def\symo{{s(\bfym;\bfyo,\theta_{mo})}}
\def\symoi{{s(\bfymi;\bfyoi,\theta_{mo})}}
\def\sc{{s_c({\bfyc};\theta_c)}}
\def\scab{{s_c({\bfyca},{\bfycb};\theta_c)}}
\def\scaba{{s_c({\bfyca};{\bfycb},\theta_c)}}
\def\scmi{{s_m(\bfymi;\theta^{(i)}_m)}}
\def\scoi{{s_o(\bfyoi;\theta^{(i)}_o)}}
\def\scmoi{{s_{mo}(\bfymi,\bfyoi;\theta^{(i)}_{mo})}}
\def\theyj{{\theta_{j}({\bar y}_j)}}
\def\theyjyk{{\theta_{jk}({\bar y}_j,{\bar y}_k)}}
\def\Iyj{{{\mathcal I}({\bar y}_j = y_j)}}
\def\Iyjk{{{\mathcal I}({\bar y}_j = y_j,{\bar y}_k = y_k)}}
\def\bfyMj{{{\bfy}_{-j}}}
\def\bfyMi{{{\bfy}_{-i}}}
\def\bfyiMj{{{\bfy}^{(i)}_{-j}}}
\def\bfyiMk{{{\bfy}^{(i)}_{-k}}}
\def\sbj{{{\bar s}_j(y_j;\bfyMj)}}
\def\sbi{{{\bar s}_i(y_i;\bfyMi)}}
\def\sbij{{{\bar s}_j(y^{(i)}_j;\bfyiMj)}}
\def\sj{{s_j(y_j)}}
\def\si{{s_i(y_i)}}
\def\sjk{{s_{jk}(y_j,y_k)}}
\def\sik{{s_{ik}(y_i,y_k)}}
\def\sij{{s_{ij}(y_i,y_j)}}
\def\qmi{{q_i(\bfymi|\bfyoi)}}
\def\pmi{{p_i(\bfymi|\bfyoi)}}
\def\mcsi{{\mathcal S}^{(i)}}
\def\mchatsi{{\hat {\mathcal S}^{(i)}}}
\def\stildej{{\tilde s}_j(\bfy)}
\def\stildeij{{\tilde s}_j(\bfy^{(i)})}



\twocolumn[

\aistatstitle{A Structured Prediction Approach for Missing Value Imputation}

\aistatsauthor{ Rahul Kidambi \And Vinod Nair \And S. Sundararajan \And S. Sathiya Keerthi }

\aistatsaddress{ Microsoft Research, India \And Microsoft Reseach, India \And Microsoft Research, India \And Microsoft, CISL, CA} ]

%\twocolumn[
%
%\aistatstitle{A Structured Prediction Approach for Missing Value Imputation}
%
%\aistatsauthor{ Anonymous Author 1 \And Anonymous Author 2 \And Anonymous Author 3 }
%
%\aistatsaddress{ Unknown Institution 1 \And Unknown Institution 2 \And Unknown Institution 3 } ]

\begin{abstract}
	Missing value imputation is an important practical problem. There is a large body of work on it, but there does not exist any work that formulates the problem in a structured output setting. Also, most applications have constraints on the imputed data, for example on the distribution associated with each variable. None of the existing imputation methods use these constraints. In this paper we propose a structured output approach for missing value imputation that also incorporates domain constraints. 	We focus on large margin models, but it is easy to extend the ideas to probabilistic models. We deal with the intractable inference step in learning via a piecewise training technique that is simple, efficient, and effective. Comparison with existing state-of-the-art and baseline imputation methods shows that our method gives significantly improved performance on the Hamming loss measure.	
  %Missing value imputation problem arises in many practical application scenarios. There exist
	%several imputation methods that address this problem. However, none of these methods provide a mechanism to
	%incorporate constraints such that the imputed values are reasonable, and possess certain desirable
	%properties specified by the user. Also, there does not exist a large margin approach addressing
	%the missing value imputation problem. Motivated by these observations, we propose a large margin structured prediction
	%learning based imputation method. Learning support vector machine for general structured
	%prediction problem is hard due to intractable inference step. Since our imputation algorithm is iterative and makes multiple calls to the learning algorithm, we need an efficient learning algorithm. To address this problem, we propose a simple and efficient piecewise training technique for learning the imputation model parameters. To the best of our knowledge, our structural SVM (SSVM) based imputation algorithm incorporating constraints is first of its kind. We conduct experiments on several benchmark datasets to compare our methods with state of the art methods. Our experimental results show that the SSVM based imputation method gives significantly improved performance on the Hamming loss measure.
\end{abstract}

\section{Introduction}

In many real world machine learning and data mining applications, data often contain missing values. While ignoring examples with missing elements is one option, this results in throwing away some valuable information such examples could provide.  There have been several approaches presented in the machine learning and statistics literature to address this problem by imputing missing values. These approaches include density estimation and Expectation-Maximization~\cite{Mar08},\cite{Gha94},\cite{Gha01},\cite{Will07}, matrix factorization~\cite{Bell07},\cite{Kor09} and conditional modeling methods~\cite{Buu07},\cite{Buu99},\cite{Tem11},\cite{Rag01},\cite{Bur10},\cite{Su11},\cite{Buu11}. 

Often many methods impute values that are unreasonable. And, the quality of imputation is assessed, for example, by comparing the distribution of observed values with the imputed values. Therefore, it will be helpful if domain constraints can be used while imputing values so that the imputed values are as per user expectation. An example constraint is: the label distribution of observed values for a variable should be same as that of the imputed values. To the best of our knowledge there does not exist any imputation method that incorporates constraints.
%\textit{One key aim of this paper is to devise imputation methods that can incorporate constraints. To the best of our knowledge there does not exist any imputation method that incorporates constraints.}.

Structured output methods have been successfully used in a number of settings. But there has been no work on formulating missing value imputation as a structured output problem.

Motivated by these observations, \textit{in this paper we develop a simple and novel approach that uses structured outputs to model variable dependencies and does imputation while satisfying domain constraints}. We develop full details for a large margin setting, but the ideas can be easily extended to probabilistic structured output models. Our learning algorithm is iterative in nature with each iteration involving a model update step and a missing value imputation step. The algorithm alternates between these two steps until there is no change in the imputed missing values and no improvement in the training objective function value.

%We model a data vector ${\bf y}$ as in a structured output problem where each variable is connected other variables. It is quite possible that there are no connections between some variables. Given the structure ${\mathcal G}$, a scoring function $\s$ is defined. The data vector for the i$^{th}$ example consists of two parts: (1) observed variables, $y_o$ and (2) missing variables, $y_m$.
%In the learning step of the algorithm, we build a large margin structured prediction model. 
Learning the structured prediction model is intractable. The main computational burden comes from solving an intractable inference problem in each iteration of the learning algorithm. A number of large margin based methods have been proposed in the literature \cite{Kom11}, \cite{Luc13}, \cite{Mes10} to address this problem via suitable approximations. Most of these methods rely on decomposing the structure into pieces or components such that inference is cheaper on the components; it is possible that variables are shared across the components. Depending on the choice of components and approximations involved, either in the formulation and/or component level inference, the training time and achievable accuracy vary.

Though they work well, we found existing approximation methods \cite{Mes10},\cite{Kom11},\cite{Luc13} to be very inefficient. \textit{To address this problem, we propose an efficient piecewise large margin learning method for learning structured prediction models.} One important aspect of the missing value imputation problem is that we need to keep in mind the noisy imputed values (obtained during the iterations) that are used in the learning step. Therefore, we present formulations where the large margin constraints are placed using only the observed values. This helps in getting robust performance. These formulations are based on the popularly known Crammer-Singer formulation (CSF) and Weston-Watkins formulation (WWF) that are used in designing large margin classifiers \cite{Cra02},\cite{Wes98}. We discuss solution methods for both these formulations. With squared hinge loss, WWF becomes differentiable, and standard unconstrained optimization techniques requiring function and gradient computations can be used. Therefore, we recommend and use piecewise WWF based method in our experiments. \textit{We found that this method is an order of magnitude faster than other state of the art large margin methods when tested on several supervised learning benchmark datasets, and there was no significant loss in accuracy}. %Therefore, we use our piecewise large margin learning method in the learning step.

The missing value imputation step involves solving a discrete optimization problem which is intractable. For this purpose, we use the popular dual decomposition method proposed by Komadakis et al \cite{Kom11b}. As emphasized earlier, imputed values should be reasonable, and should have certain desirable properties as expressed by the user. Furthermore, it is well known in the semi-supervised learning literature \cite{Joa99},\cite{Dhi12},\cite{Cha13} that constraints play an important role in getting good accuracy. As we shall see, incorporating constraints is easier in our approach, and also, they give significant boost to the performance. Recently, Chang et al.\ \cite{Cha13} proposed a dual decomposition method that can handle constraints. We use their method for imputing the missing values subject to constraints. \textit{Overall, the proposed large margin structured prediction model based imputation method that can incorporate domain knowledge via constraints is an important contribution in addressing the missing value imputation problem}.

The paper is organized as follows. We present related work in Section 2. In Section 3, we present our approach in detail. In Section 4, we demonstrate the usefulness of the proposed imputation method compared to state of the art methods on several benchmark datasets. Experimental results show that our method outperforms these methods on the Hamming loss measure. Discussion and Future work are presented in Section 5. We conclude with Section 6.

\section{Related Work}
One important class of missing value imputation algorithms is based on density estimation. A standard approach is to learn the joint distribution of the data as a mixture model by treating missing values as latent variables, and using the Expectation Maximization (EM) algorithm \cite{Mar08},\cite{Gha94} or Variational Bayes EM \cite{Gha01},\cite{Will07},\cite{Win05}. Then use the learned model to infer the the posterior distribution over the missing values given the observed values. The drawback is that accurately modeling the joint distribution of high-dimensional data is a hard problem, and it is not strictly necessary if the goal is to only predict the missing values. By adopting a discriminative approach, we avoid the difficulty of joint distribution learning altogether.

Algorithms based on matrix factorization and neighborhood models are popular in collaborative filtering (CF) for predicting missing entries of a ratings matrix \cite{Bell07},\cite{Kor09}. The missing value imputation setting differs from CF in important ways: 1) the dataset can contain mixed variable types (real, categorical, binary, etc.), so matrix factorization based algorithms need to be modified to such mixed type data, and 2) CF applications have much higher missing percentages (e.g. 99\%), and 3) evaluation is focused more on predicting a few relevant missing entries rather than \emph{all} of them.

Several imputation algorithms in the statistics literature use a ``pseudo-Gibbs'' sampling procedure that alternates between 1) sampling the missing values and 2) sampling the parameters of a predictor for each variable that uses all other variables as inputs \cite{Buu07},\cite{Buu99},\cite{Tem11},\cite{Rag01},\cite{Bur10},\cite{Su11},\cite{Buu11}. Multiple Imputations using Chained Equations (MICE) \cite{Buu11} is a popular example of this approach. It is well-known that MICE and its variants have no convergence guarantees, and can even be shown to diverge in some special cases \cite{Liu10},\cite{Li12}. In contrast, our learning algorithm is a convex optimization problem, and our imputation algorithm allows for constraints in the optimization of the missing values, while MICE does not. Empirically, MICE is indeed one of the best imputation methods, and hence we evaluate our method against it to demonstrate value.

\section{Our Approach}
We present details of our large margin structured prediction learning approach for solving the missing value imputation problem. The proposed method is iterative in nature, and involves two key steps. (1) Learning step (Section 3.2): learn a structured prediction model given the {\it full} data matrix. (2) Imputation Step (Section 3.3): impute missing values subject to constraints given the model parameters. See Algorithm~\ref{Algorithm}. As we shall see, our large margin learning approach to learn the model parameters is different from conventional methods.  After giving the details of the learning step, we discuss a dual decomposition based method to impute the missing values subject to distribution constraints \cite{Cha13}.

\begin{algorithm2e}
\label{Algorithm}
\caption{Large Margin Structured Prediction based Imputation Method}
Choose $\bfym$ (e.g., fill-in using the mode value for each column), {\it tol} (e.g., $10^{-4}$), {\it iter} (e.g., $25$)\;
\For{$r=1 \ldots, iter$}{
1. {\bf Learning Step}: Compute model parameters ($\theta$) by solving the learning problem (\ref{wwfp})\;
2. {\bf Imputation Step}: Solve the imputation problem (\ref{scrimpstep}). Update $\bfym$\;
3. Exit if the relative improvement in the objective function value is less than {\it tol}\;
}
\end{algorithm2e}
%We start with the problem statement and relevant background on: (a) structured prediction (data) model based on Markov Random Fields (MRF) that we use, and (b) structure decomposition and associated scoring functions needed for piecewise learning. This is followed by a simple and efficient piecewise large margin formulation, and explain how we learn the model parameters by solving an associated optimization problem. Then, we present an imputation algorithm that imputes the missing values subject to the constraints; this algorithm is based on the dual decomposition method (give Komadakis et al and Kai-Wei et al).
\subsection{Background}
\noindent{\bf Notations} Let $\bfy$ denote the $K$-dimensional label vector with $j^{th}$ variable $y_j \in {\mathcal Y}_j$, where ${\mathcal Y}_j$ is the label space. Let $\bfyi$ denote the $i^{th}$ example where $i \in \{1,\ldots,n\}$; $\bfyoi$ and $\bfymi$ denote the observed and missing variables in the $i^{th}$ example. Let ${\bf Y}$ denote the data matrix with $\bfyi$ being the $i^{th}$ row. We use $|\cdot|$ to indicate the cardinality of a set. 

\noindent{\bf Problem Statement} Given the data matrix ${\bf Y}$, the goal is to fill-in (impute) values for the missing entries, \textit{i.e.,} given $\{\bfyoi: i=1,\ldots,n\}$, get values for $\{\bfymi: i=1,\ldots,n\}$. This paper concerns working with discrete variables. Continuous variables can be handled by applying a suitable discretization technique.

%\subsection{Background}
\noindent{\bf Structure and Scoring Function Model} We model the data as in structured output problems. In these problems, the random variables constituting $\bfy$ are connected as defined by a network ${\mathcal G}$. Given the structure ${\mathcal G}$, a scoring function returns a value for any given assignment to the variables. In our study, we consider structures with scoring function defined as:
\begin{equation}
\s = \sum_{c \in {\mathcal C}} \sc
\label{scfn}
\end{equation}
where $c$ and ${\mathcal C}$ denote $c^{th}$ component in a collection of components ${\mathcal C}$ in ${\mathcal G}$; $\bfyc$ and $\theta_c$ denote the $c^{th}$ component variables and associated model parameters. Let us assume that $\sc$ takes the form:
\begin{equation}
\sc = \sum_{\alpha} \theta_{c,\alpha} \phi_{\alpha}(\bfyc)
\label{scorec}
\end{equation}
where $\phi_{\alpha}(\cdot)$ denote the sufficient statistics for $\bfyc$. One example is: $\phi_{\alpha}(\bfyc) = {\mathcal I}(\bfyc) = \prod_{y_j \in \bfyc} {\mathcal I}({\bar y}_j = y_j)$ where ${\mathcal I}(\cdot)$ is the indicator function taking value $1$ when the argument is true, and $0$ otherwise, and $\alpha = {\brbfyc} \in {\mathcal Y}_c$. The approach presented in this paper is quite generic; to make explanations easier and concrete, we illustrate through pairwise Markov Random Field (MRF).

\noindent{\bf Pairwise MRF Example} Consider a network ${\mathcal G}$ where all the variables are connected to each other, with the scoring function $\s$ defined as:
\begin{equation}
\s = \sum_{j=1}^K \sj + \sum_{j=1}^K \sum_{k = j+1}^K \sjk
\label{scmrfn}
\end{equation}
where $\sj$ and $\sjk$ denote the scores for node $j$ and edge $(j,k)$ respectively. In linear form, they are given by: $\sj = \sum_{{\bar y}_j \in {\mathcal Y}_j} \theyj \Iyj$, and $\sjk = \sum_{{\bar y}_j \in {\mathcal Y}_j} \sum_{{\bar y}_k \in {\mathcal Y}_k} \theyjyk \Iyjk$. 

\noindent{\bf Imputation} Given the model parameters, imputation of missing values can be done via solving the following discrete (variables) optimization problem:
\begin{equation}
\bfyms = \argmax_{\bfym} \symo
\label{infpbm}
\end{equation}
where $\theta_{mo}$ denotes the model parameters involved in connecting missing variables with themselves and observed variables. Note that the connections involving only the observed variables do not play a role. The main issue in the structured prediction setting is that the cardinality of the label space of $\bfym$ can be exponential. Therefore, solving the inference problem is intractable (except for simple structures such as trees).
Komadakis et al \cite{Kom11b} proposed a dual decomposition method that can be used to solve (\ref{infpbm}).

Given the model, scoring function and basic imputation step, we show how a large margin structured prediction model with the scoring function (\ref{scmrfn}) can be built into the \textit{learning step} of the algorithm.

\subsection{Learning Step}

\noindent{\bf Large Margin Learning Approach} There are four key elements that need to be defined in learning large margin structured prediction models: (1) weight regularization, (2) an error function $\Delta(\bfy,\bfy^{(i)})$ that specifies the margin that we would like to impose between the true label assignment $\bfy^{(i)}$ and any other assignment ${\bar \bfy} \in {\mathcal Y}^{(i)}$ for the $i^{th}$ example; ${\mathcal Y}^{(i)}$ denotes the label space, (3) large margin constraints, and (4) loss function (e.g., hinge loss, squared hinge loss) defined using slack variables that appear in the constraints. Using these elements, an objective function is defined as a linear combination of the weight regularization and loss function terms, and optimized for the model parameters (weights). In our study, we use the L2 norm for weight regularization, and the Hamming distance for the error function. We will discuss our choice of large margin constraints and loss function shortly.

\noindent{\bf A Simple Idea} Let us assume that $\{\bfymi: i=1,\ldots,n\}$ is available; for example, we can fill-in the missing values with the mode value of each variable, or sample according to the distribution over the observed values. Then, we can consider the learning problem as learning a structured prediction model in supervised setting, where $(\bfymi,\bfyoi)$ is treated as an input-output example pair. Then, the learning problem can be posed as:

CSF:
\begin{equation}
%\textrm{CSF:} \quad
\min_{\bf \theta} \frac{||\theta||^2}{2} + \lambda \sum_{i=1}^n \max(0,\psi^{(i)}).
\label{baseform}
\end{equation}
Let ${\hat \bfy}^{(i)}_o = \argmax_{{\bar {\bf y}}_o \in {\mathcal Y}^{(i)}_o} s^{(i)}({\bar \bfy}_o;\bfymi) + \Delta(\bfy^{(i)}_o,{\bar \bfy}_o)$, $\psi^{(i)} = s^{(i)}({\hat \bfy}^{(i)}_o;\bfymi)  + \Delta(\bfyoi,{\hat \bfy}^{(i)}_o) - s^{(i)}(\bfyoi;\bfymi)$, $\lambda$ is a regularization constant, ${\mathcal Y}^{(i)}_o=\prod_{j: j \in I^{(i)}_o} {\mathcal Y}_j$, $I^{(i)}_o$ is the index set of the observed variables, and ${\mathcal Y}_j$ is the label space of the $j^{th}$ variable; $s^{(i)}(\cdot)$ denotes the score associated with the $i^{th}$ example. This primal formulation arises from using the large margin constraints $s^{(i)}(\bfyoi;\bfymi) - s^{(i)}({\bar \bfy}_o;\bfymi) \ge \Delta(\bfyoi,{\bar \bfy}_o) - \psi^{(i)}, \forall i,{\bar \bfy}_o \in {\mathcal Y}^{(i)}_o$ with one slack variable per example, and optimizing over the model parameters and slack variables using hinge loss as the loss function; this formulation is also known as the Crammer-Singer Formulation (CSF) \cite{Cra02}.

We make the following observations. (1) The number of target variables that appear in each example is different. (2) The total number of input and output variables is a constant $K$; this is usually not the case, in structured output problems. (For example, consider sequence or tree labeling problems; though the input feature dimension can be same, the number of outputs (i.e., nodes in a sequence or tree) needs not be same for all the sequences or trees.) (3) We have suppressed the dependency of the model parameters in the scoring functions. The most important thing to note is that not all the model parameters appear in every example, and, some model parameters are shared across the examples depending on the patterns of missing variables. Some of these characteristics are specific to learning a structured prediction model for the missing value imputation problem compared to conventional supervised learning of structured prediction models.

One way to solve the optimization problem (\ref{baseform}) is to use a stochastic sub-gradient algorithm~\cite{Rat07}, \cite{Kiw83}, \cite{Luc13} where the sub-gradient is computed for a randomly picked example in each iteration, and the model weights are updated. This step involves solving a modified inference problem with a decomposable error function $\Delta(\cdot)$ added to the scoring function in (\ref{infpbm}). Solving this inference problem is intractable except for simple structures such as trees or when $|\bfymi|$ is small and finding the maximum is easy. To address this problem several efficient methods have been proposed in the literature \cite{Kom11b},\cite{Wai03}. One approach that is relevant to this work is piecewise training \cite{Sut09} where the structure is decomposed into pieces or components such that inference in each of the components is tractable; this helps in reducing the training complexity. While the existing piecewise large margin training methods \cite{Mes10},\cite{Kom11} can be used to solve (\ref{baseform}), we need a much more efficient learning algorithm. This is because our imputation method is iterative in nature, and the learning problem needs to be solved repeatedly. Our experiments show that the speeds offered by the existing methods are not good enough. To address this problem, we propose a simple piecewise large margin formulation, which we will also demonstrate to be very effective; and the resulting optimization problem is differentiable, and standard unconstrained optimization algorithms such as L-BFGS~\footnote{http://www.di.ens.fr/~mschmidt/Software/minFunc.html} can be used. Before describing the final method, we first present a piecewise version of (\ref{baseform}).

\noindent{\bf Piecewise Large Margin Training} The basic idea behind our formulation can be explained as the following steps. (1) For each example $i$, we construct multiple pairs of observed variables $\bfyoi$ by partitioning $\bfyoi$ into two parts in multiple ways. Let $\{{\bf u}^{(i)}_{q,o}: q=1,\ldots,Q_i\}$ denote choices for the first part, and let ${\bar {\bf u}}^{(i)}_{q,o} = \bfyoi \setminus {\bf u}^{(i)}_{q,o}, \forall i,q$. (2) Given these sets, we construct input-output example pairs: $({\tilde {\bf u}}^{(i)}_{q,o},{\bf u}^{(i)}_{q,o}), \forall i, q$ where ${\tilde {\bf u}}^{(i)}_{q,o} = [{\bar {\bf u}}^{(i)}_{q,o}\:\bfymi]$. In other words, we form the input by appending one part of the observed variables ${\bar {\bf u}}^{(i)}_{q,o}$ with $\bfymi$, and the other part ${\bf u}^{(i)}_{q,o}$ forms the output. Thus, each example $(i)$ is repeated multiple times ($Q_i$) with different input-output pairs. The important aspect of such subset formation is that $|{\bf u}^{(i)}_{q,o}|$ is small so that finding the maximum is easy (e.g., via enumeration with one or two variables). (3) Using this constructed dataset $\{({\tilde {\bf u}}^{(i)}_{q,o},{\bf u}^{(i)}_{q,o}), \forall i, q\}$, we form the constraints for each example. (We show how this dataset is constructed for a pairwise MRF below.) This results in the following optimization problem in Crammer-Singer Formulation-Piecewise (CSFP).

CSFP:
\begin{equation}
\min_{\bf \theta} \frac{||\theta||^2}{2} + \lambda \sum_{i=1}^n \sum_{q=1}^{Q_i}\max(0,\psi^{(i)}_q),
\label{csfp1}
\end{equation}
Let ${\hat {\bf u}}^{(i)}_{q,o} = \argmax_{{\bar {\bf u}}_{q,o} \in {\mathcal Y}^{(i)}_{q,o}} s^{(i)}({\bar {\bf u}}_{q,o};{\tilde {\bf u}}^{(i)}_{q,o}) + \Delta({\bf u}^{(i)}_{q,o},{\bar {\bf u}}_{q,o})$, $\psi^{(i)}_q = s^{(i)}({\hat {\bf u}}^{(i)}_{q,o};{\tilde {\bf u}}^{(i)}_{q,o})  + \Delta({\bf u}^{(i)}_{q,o},{\hat {\bf u}}^{(i)}_{q,o}) - s^{(i)}({\bf u}^{(i)}_{q,o};{\tilde {\bf u}}^{(i)}_{q,o})$, and ${\mathcal Y}^{(i)}_{q,o}=\prod_{j: j \in I^{(i)}_{q,o}} {\mathcal Y}_j$; $I^{(i)}_{q,o}$ is the index set of the observed variables ${\bf u}^{(i)}_{q,o}$. As earlier, this primal formulation arises from using the large margin constraints $s^{(i)}({\bf u}^{(i)}_{q,o};{\tilde {\bf u}}^{(i)}_{q,o}) - s^{(i)}({\bar {\bf u}}_{q,o};{\tilde {\bf u}}^{(i)}_{q,o}) \ge \Delta({\bf u}^{(i)}_{q,o},{\bar {\bf u}}_{q,o}) - \psi^{(i)}_q, \forall i,{\bar {\bf u}}_{q,o} \in {\mathcal Y}^{(i)}_{q,o}$ with one slack variable per piece per example, and optimizing over the model parameters and slack variables using hinge loss as the loss function.
%denoting ${\tilde \bfy}^{(i)}_o = \argmax_{{\bar {\bf y}}_o \in {\mathcal Y}^{(i)}_o} s^{(i)}({\bar \bfy}_o;\bfymi) + \Delta(\bfy^{(i)}_o,{\bar \bfy}_o)$, $\psi^{(i)} = s^{(i)}({\tilde \bfy}^{(i)}_o;\bfymi)  + \Delta(\bfyoi,{\tilde \bfy}^{(i)}_o) - s^{(i)}(\bfy^{(i)};\bfymi)$, $C$ is a regularization constant, ${\mathcal Y}^{(i)}_o=\prod_{j: j \in I^{(i)}_o} {\mathcal Y}_j$, $I^{(i)}_o$ is the index set of the observed variables, and ${\mathcal Y}_j$ is the label space of the $j^{th}$ variable. This primal formulation arises from using the large margin constraints $s^{(i)}(\bfyoi;\bfymi) - s^{(i)}({\bar \bfy}_o;\bfymi) \ge \Delta(\bfyoi,{\bar \bfy}_o) - \psi^{(i)}, \forall i,{\bar \bfy}_o \in {\mathcal Y}^{(i)}_o$ with one slack variable per example, and optimizing over the model parameters and slack variables using hinge loss as the loss function;

The following observations can be made. (1) The optimization problem (\ref{csfp1}) can be solved using stochastic sub-gradient algorithm. (2) The number of constraints used in (\ref{csfp1}) is dependent on $Q_i$ and how the partitions of $\bfyoi$ are defined. (3) Since the constraints used in (\ref{baseform}) and (\ref{csfp1}) are different, the solutions will be different. (Our experimental results in supervised setting show that there is no significant loss in accuracy; typically, the loss is within 1-1.5\% as observed in fully supervised multi-label classification datasets.) But, the important point is that we gain by an order of magnitude in speed. (4) The above problem can be solved using an efficient multi-class dual coordinate descent method \cite{Kee08} as the cardinality $|{\bf u}^{(i)}_{q,o}|$ is small. (5) While the optimization problem (\ref{csfp1}) is Crammer-Singer formulation based, we can have another popular formulation known as Weston-Watkins formulation (WWF) with squared hinge loss. This formulation with squared hinge loss is differentiable; therefore, any standard unconstrained optimization technique can be used. Our experimental results show that this method is very efficient. We present this formulation next.

\noindent{\bf Weston-Watkins Formulation (WWF)} In CSF, the number of slack variables involved per piece per example is $1$. On the other hand, one slack variable is associated with each possible label assignment ${\bar {\bf u}}_{q,o} \in {\mathcal Y}^{(i)}_{q,o}, \forall i,q$. Then, the optimization problem in the Weston-Watkins Formulation - Piecewise (WWFP) is given by:

WWFP:
\begin{equation}
\min_{\bf \theta} \frac{||\theta||^2}{2} + \frac{\lambda}{2} \sum_{i=1}^n \sum_{q=1}^{Q_i} \sum_{{\bar {\bf u}}_{q,o} \ne {\bf u}^{(i)}_{q,o}} \max(0,\psi^{(i)}({\bar {\bf u}}_{q,o}))^2,
\label{wwfp}
\end{equation}
where $\psi^{(i)}({\bar {\bf u}}_{q,o}) = s^{(i)}({\bar {\bf u}}_{q,o};{\tilde {\bf u}}^{(i)}_{q,o})  + \Delta({{\bf u}}^{(i)}_{q,o},{\bar {\bf u}}_{q,o}) - s^{(i)}({\bf u}^{(i)}_{q,o};{\tilde {\bf u}}^{(i)}_{q,o})$. Comparing (6) and (7), we note that (7) has an extra, third summation at the innermost level; though this leads to more loss terms, this is the key to getting a differentiable objective function, with associated efficiency. In our experiments on the missing value imputation problem, we used WWFP since fast optimization techniques such as L-BFGS can be used, leading to great improvements in speed. Also, it has been found that there is no evidence of significant difference in generalization performance between CSFP and WWFP.

\noindent{\bf Partitioning Examples} As mentioned earlier, partitions can be created in several ways. We give two examples. One simple example is to take $|{\bf u}^{(i)}_{q,o}|=1$, and consider each variable in $\bfyoi$ as ${\bf u}^{(i)}_{q,o}$ as we vary $q$; for this case, we have $Q_i = |\bfyoi|$. Another example is to take every pair of variables in $\bfyoi$ as ${\bf u}^{(i)}_{q,o}$. In this case, $Q_i  = \frac{|\bfyoi|(|\bfyoi|-1)}{2}$. Note that it is not necessary to restrict to only one type of partitioning (\textit{i.e.,} single variable or pairwise); we can have both types in (\ref{wwfp}). For the purpose of illustration, let us consider the single variable type partitioning and see how the scoring function looks like in the pairwise MRF model. In pairwise MRF, the scoring function $\s$ can be decomposed as: $\s = \sum_j \sbij$ where
\begin{equation}
\sbij = \sj + \sum_{k \ne j} \frac{\sjk}{2}
\label{score}
\end{equation}
This decomposition has the interpretation of decomposing the structure into $K$ spanning trees with $\sbij$ representing the score of $j^{th}$ spanning tree with the $j^{th}$ variable as the root node. This is the scoring function that we use in (\ref{wwfp}) with $u^{(i)}_{q,o} = y^{(i)}_j: j\in I^{(i)}_{q,o}, q=1,\ldots,|\bfyoi|$. This completes the specification of all quantities in (\ref{wwfp}). It is worth noting that the model parameters are shared across the trees as edges are shared across the trees.

%\noindent{\bf Supervised Learning Setting} An input feature vector ${\bf x}$ is associated with the label vector $\bfy$, and the scoring function taking ${\bf x}$ into account can be defined as: $\sc = \sum_\alpha \theta_{c,\alpha} \cdot \phi_\alpha({\bf x},\bfyc)$ where $\cdot$ denotes the dot product of the weight and feature vectors. Putting together, the overall score is: $\s = \theta \cdot \phi({\bf x},{\bf y})$ where we have collated all the component weight and feature vectors into $\theta$ and $\phi(\cdot)$ respectively. Note that we have omitted the dependency of the left hand side on $\bfx$, for notational simplicity. An example of (\ref{scmrfn}) is: $\s = \sum_{j=1}^K \theta_{y_j} \cdot {\bf x} + \sum_{k>j} \theta_{y_j,y_k} \cdot {\bf x}$ where the individual dot products define the node and edge potentials of the pairwise MRF. For example, such a model can be used to solve a multi-label classification problem that commonly arises in web document and image classification contexts.

\subsection{Imputation Step}
While (\ref{infpbm}) is useful to impute missing values for each row independently, our goal is to impute $\{\bfymi: i=1,\ldots,n\}$ jointly. Joint imputation becomes a necessity when constraints that involve multiple rows are specified, as discussed below.

\noindent{\bf Constraints} We give two examples of constraints that we used in our experiments. These constraints are specified as distributions over individual and pair of variables.

(1) {\bf Label Distribution}: Let $p_j$ denote the label distribution over ${\mathcal Y}_j$. Constraints of this form are useful to quantitatively specify one consistency check that is done manually by visually looking at the distribution of observed and imputed values for each variable; this visual consistency check is provided with several practically used statistical methods (e.g., MICE).

(2) {\bf Pairwise Distribution}: Let $q_{j,k}$ denote the pairwise distribution over ${\mathcal Y}_j \times {\mathcal Y}_k$. These constraints are $2$-d extension of label distribution constraints, and are useful to capture co-occurrence statistics. 

Given the observed data, we can compute $p_j$ and $q_{j,k}$ using the observed values; and, constrain the imputed values to have the same distributions. When multiplied by the number of examples, these distributions specify the fraction of examples (${\tilde n}_j$, ${\tilde n}_{j,k}$) in which the labels and pairs of labels occur. Let ${\mathcal C}$ denote these constraints. Since these constraints span across the rows, imputation cannot be done independently via (\ref{infpbm}).

\noindent{\bf Optimization Problem} The imputation step involves solving the following optimization problem.
\begin{equation}
\max_{{\bf Y}_m} \sum_{i=1}^n s^{(i)}(\bfymi;\bfyoi,\theta_{mo}) \:\: \rm{s.t.}\:\: {\mathcal C}
\label{scrimpstep}
\end{equation}
where ${\bf Y}_m = \{\bfymi:i=1,\ldots,n\}$. This problem can be solved using a joint inference method proposed by Chang et al \cite{Cha13}. They extended the dual decomposition method \cite{Kom11b} to handle constraints. It is an iterative method that alternates between imputing the missing values and finding the dual parameters associated with the constraints. Due to constraints and non-differentiability reasons, a projected sub-gradient method is used to solve these sub-problems in the alternating optimization steps.

\section{Experiments}

We present our experimental results from two sets of experiments. The goal of the first set of experiments is to provide evidence that the proposed WWFP based piecewise training method is an order of magnitude faster than state of the art methods; we considered a multi-label supervised learning problem for this study. We found this speed-up gain is very important since the overall imputation method (algorithm~\ref{Algorithm}) typically takes $10$-$15$ iterations. The main focus is the second set of experiments where we show that the proposed large margin based imputation method outperforms state of the art imputation methods on several benchmark datasets.

\subsection{Supervised Learning Experiments}
We considered supervised multi-label classification problem, and we used two benchmark datasets: (1) \textsc{Medical} and \textsc{scene}. These datasets are available at~\url{http://mulan.sourceforge.net/datasets.html}. For the \textsc{Medical} dataset, we considered top 10 output variables selected as per the class proportion; outputs with high class imbalance were dropped. We compared our WWFP method with three state of the art large margin structured prediction model learning methods; the first method is dual decomposition based learning (DDL) method~\cite{Kom11}, and the second method~\cite{Luc13},\cite{Rat07} is a primal stochastic sub-gradient (SSG) method. We use test set Hamming loss as the measure for comparing the performance. On the \textsc{Medical} dataset, all the methods (averaged over 2 splits) gave a test error of 9.36\%, and the training time taken by WWFP, DDL and SSG were 54.6, 349.5 and 4662.6 seconds respectively. On the \textsc{Scene} dataset, the test errors obtained were 11.37\%, 10.62\% and 9.89\%, and the training times were 7.74, 147.7, 6073 seconds. These results clearly show that the proposed method achieves significant training time speed-up without significant loss in accuracy.

\subsection{Missing Value Imputation Experiments}
\subsubsection{Experimental Setup}

\noindent{\bf Datasets} We used four categorical datasets from the UCI repository (\url{http://archive.ics.uci.edu/ml/}). Details of these datasets are listed in table \ref{t1}. Note that none of these datasets contain missing values originally. We introduce missing values completely at random such that a certain fraction of each column of the dataset is missing. In our experiment, we consider 3 missing percentages, i.e., we allow 10/30/50\% of the dataset to be missing.
\begin{table}
\centering
%\resizebox{\columnwidth}{!}{
\begin{tabular}{|c|c|c|}
\hline
dataset & \# variables & type\\
\hline
flare & $9$ & multinomial\\
\hline
spect & $22$ & binomial\\
\hline
mushroom & $10$ & multinomial\\
\hline
yeast & $14$ & binomial\\
\hline
\end{tabular}
%}
\caption{Details of various datasets used in the missing value imputation experiments}
\label{t1}
\end{table}

\noindent{\bf Measure for Comparison} We use Hamming loss on the imputed values. We get one table at the end of the optimization from our methods, \textsc{Mode} and \textsc{BF}, and we compare this output with the ground truth table to compute the Hamming loss. \textsc{MICE} and \textsc{MM} are sampling based methods; therefore, these methods can generate multiple imputations. We generate 100 samples, and then compute the expected Hamming loss by averaging over the Hamming loss obtained using these 100 tables.

\noindent{\bf Methods for Comparison} We compare our large margin method with four popular methods in the statistics literature. They are:
\begin{enumerate}
\item \textsc{Mode} (MO): Fills up all missing values in a column with the mode of the observed values in the column.
\item \textsc{Mixture Model} (MM): A model in which each mixture component is a product of univariate distributions over each variable. Inference and learning performed using variational Bayes inference\cite{Win05}.
\item \textsc{Mice} (MICE): State of the art imputation algorithm\cite{Buu11}. Involves an iterative procedure of training predictors over each output variable (by minimizing the logistic loss), keeping the other variables as inputs, and sampling the predictor's output distributions over the states of each variable to generate the imputations of the dataset. Note that both parameter learning and imputation is carried out through sampling the posterior distribution over the model weights and states of the variables respectively.
\item \textsc{Backfitting} (BF): A deterministic version of \textsc{Mice}, where instead of the sampling step, we select the optimal parameters obtained from solving the learning (optimization) problem using hinge loss, and the imputed values are obtained by solving the individual inference problem.
\item \textsc{WWFP-WO} Our proposed method without constraints. Note that imputations are done independently for each row by solving (\ref{infpbm}).
\item \textsc{WWFP-WC} Our proposed method with label and pairwise distribution constraints. In this case, joint inference is needed, as explained in Section 3.3.
\end{enumerate}

\noindent{\bf Implementation Details} To set the regularization constant $\lambda$, we perform a cross validation step as follows. Given the missing value data, we build a model for each $\lambda$. Using the model built, for each column, we predict the observed variables in that column using values for the remaining variables picked from the ground truth data. We average the zero-loss computed on each column. Finally, we choose the constant that gives the minimum loss. To make a fair comparison, we used this approach uniformly for all the methods. We are exploring other cross validation approaches for the missing value data scenario. To evaluate the performance, we create six splits of each dataset, and run all the imputation algorithms on all the splits. The results are averaged across the splits of each dataset and presented.  
%One such method is: (1) construct a dataset by making some observed values missing from the input data with missing values, (2) build a model using this dataset for different $\lambda$ values, and (3) evaluate the performance on the observed values that were made missing. 

\begin{table*}
\setlength{\extrarowheight}{6pt}
\centering
\resizebox{\linewidth}{!}{
\begin{tabular}{|c|c|c|c|c|c|c|c|}
\hline
dataset & \textsc{MO} & \textsc{MM} & \textsc{MICE} & \textsc{BF} & \textsc{WWFP-WO} & \textsc{WWFP-WC}\\
\hline
FLARE (10\%) & $34.15\pm1.74$ & $29.59\pm0.76$ & $25.48\pm0.9$ & $20.04\pm1.27$ & $20.75\pm2.07$ & {$\bf19.87\pm1.77$}\\
\hline
SPECT (10\%) & $30.23\pm2.95$ & $31.39\pm1.47$ & $23.76\pm1.45$ & $18.76\pm2.01$ & $17.57\pm1.58$ & {$\bf15.29\pm1.7$}\\
\hline
MUSHROOM (10\%) & $45.39\pm0.88$ & $35.5\pm1.69$ & $27.78\pm0.50$ & {$\bf 25.51\pm0.63$} & $30.95\pm0.97$ & $25.54\pm0.52$\\
\hline
YEAST (10\%) & $23.26\pm0.74$ & $23.99\pm1.73$ & $9.62\pm0.83$ & {$\bf 6.59\pm0.82$} & $9.20\pm0.85$ & $7.00\pm1.48$\\
\hline
\hline
FLARE (30\%) & $34.25\pm0.5$ & $32.89\pm1.12$ & $28.51\pm0.29$ & $23.71\pm0.6$ & $24.86\pm1.24$ & {$\bf22.48\pm0.76$}\\
\hline
SPECT (30\%) & $30.38\pm2.0$ & $31.22\pm0.9$ & $25.8\pm0.67$ & $21.98\pm1.09$ & $19.35\pm1.09$ & {$\bf17.77\pm1.01$}\\
\hline
MUSHROOM (30\%) & $45.57\pm0.62$ & $40.62\pm1.93$ & $30.11\pm4.96$ & $28.56\pm0.48$ & $34.25\pm1.01$ & {$\bf28.27\pm0.57$}\\
\hline
YEAST (30\%) & $23.13\pm0.74$ & $27.01\pm0.55$ & $15.33\pm0.42$ & {$\bf 11.54\pm0.8$} & $13.75\pm0.91$ & $11.7\pm0.63$\\
\hline
\hline
FLARE (50\%) & $34.67\pm0.22$ & $39.3\pm0.9$ & $32.24\pm0.53$ & $27.34\pm0.71$ & $28.7\pm1.12$ & {$\bf25.76\pm0.69$}\\
\hline
SPECT (50\%) & $30.93\pm1.01$ & $33.87\pm0.75$ & $28.86\pm0.29$ & $25.11\pm1.12$ & $28.28\pm1.9$ & {$\bf19.25\pm0.62$}\\
\hline
MUSHROOM (50\%) & $45.59\pm0.46$ & $50.36\pm1.68$ & $38.05\pm0.41$  & $33.96\pm0.53$ & $38.80\pm0.54$ & {$\bf32.7\pm0.36$}\\
\hline
YEAST (50\%) & $23.12\pm0.51$ & $31.66\pm0.36$ & $20.91\pm0.124$  & $15.87\pm0.61$ & $17.77\pm0.8$ & {$\bf15.7\pm0.38$}\\
\hline
\end{tabular}
}
\caption{Hamming Loss (\%) for different percentage of missing data indicated in brackets.}
\label{tbl2}
\end{table*}

\subsubsection{Experimental Results}
Experimental results on the benchmark datasets are shown in Tables~{\ref{tbl2}}. The best performance numbers are highlighted in bold. From these tables, the following observations can be made. First, on comparing the last two columns corresponding to our methods without and with constraints clearly demonstrate that significant performance improvement can be achieved by incorporating constraints. Second, the best performance is achieved by \textsc{WWFP-WC} in $9$ out of $12$ cases. In particular, this method starts dominating as the percentage of missingness ($pm$) increases. This is expected because as $pm$ increases, the amount of information available in the data decreases; therefore, additional information such as constraints help in getting improved performance. Similar observation has been made in the semi-supervised learning literature~\cite{Cha13},\cite{Dhi12}.

The performance of \textsc{BF} is close to our method in some cases; and, \textsc{BF} is the second best method. This method has two advantages: (1) it is easy to implement, and (2) is faster compared to our method. When the accuracy performance is of great importance, our method is preferred. The popular \textsc{MICE} method comes next followed by \textsc{MM}. Worst performance was achieved by the simple baseline method \textsc{MO}. This is expected since there is no information used from other variables; also, the estimates are crude.

\section{Discussion and Future Work}

Probabilistic methods such as \textsc{MICE} and \textsc{MM} have the advantage that multiple imputations can be made. This is useful to capture uncertainty in imputations. On doing some detailed analysis, we found that the performance achieved by \textsc{MICE} using the mode computed from the generated samples was found to be good, and very close to \textsc{BF}. However, as the results show, the expected loss is worse; this is because significant mass was present on other samples with high Hamming loss. Similar observation was made with the \textsc{MM} method as well. These observations suggest that there is a need to come out with probabilistic imputation methods that give good performance in expectation. In this context, it is worthwhile to modify our method to make it a probabilistic model. The piecewise learning ideas developed in [3] can be suitably adapted for this purpose.

We observed in our experiments that our method converged in $10$-$15$ iterations. Although we never faced any convergence issue, there is no formal convergence proof to our method. Our method can be improved in the imputation step by changing the optimization problem. For example, instead of solving (\ref{scrimpstep}), we can solve the following optimization problem:
\begin{equation}
\max_{{\bf Y}_m} \sum_{i=1}^n \sum_{q=1}^{Q_i} \sum_{{\bar {\bf u}}_{q,o} \ne {\bf u}^{(i)}_{q,o}} \max(0,\psi^{(i)}({\bar {\bf u}}_{q,o}))^2 \:\: \rm{s.t.}\:\: {\mathcal C}
\label{impstep}
\end{equation}
This formulation has the advantage that the objective function used in the imputation step is also used in the learning step. Therefore, we would expect the objective function to decrease in every step. With bounded objective function value, the method will converge. However, the imputation step becomes complex with the WWFP formulation because of the squared hinge loss term. This term produces higher order potentials in the form of products of scoring functions; for example, multiplying one pairwise scoring function with another results in higher order potentials. Although dual decomposition can be used to solve the problem, the complexity increases and, the overall time taken by the method goes up. On the other hand, higher order potentials would not appear when hinge loss is used; but, the learning step becomes slower due to the non-differentiable nature of the problem. Therefore, there is a trade-off involved in getting overall improvement in speed. This aspect of the problem requires further study.

The proposed method can also be used to solve large margin semi-supervised learning problems for structured outputs. For example, our method can be extended to solve multi-label classification problems, when only partially labeled information is available. The extension is straight-forward as the scoring functions can be appropriately modified to incorporate input features.

While we incorporated distributions based constraints, a detailed study incorporating other types of constraints in different missing value scenarios (e.g., when data is not missing completely at random) is an interesting future work. Finally, though our method can be applied when continuous variables are discretized, coming up with an imputation method that handles mixed data types (e.g., real, categorical and ordinal) directly is another important research problem. 

\section{Conclusion}
We proposed a large margin structured prediction modeling based imputation method to solve the missing value problem. We showed how constraints can be incorporated while imputing the missing values. This is a very powerful feature since consistency checks that are often done to assess the quality of imputation methods can be quantitatively integrated while imputing the missing values. As our experimental results show constraints help in getting significantly improved performance.

\bibliographystyle{plain}
\bibliography{PaperBib}

\begin{thebibliography}{10}

\bibitem{Bell07}
R.M. Bell and Y.~Koren.
\newblock Scalable collaborative filtering with jointly derived neighborhood
  interpolation weights.
\newblock In {\em ICDM}, pages 43--52, 2007.

\bibitem{Bur10}
L.~F. Burgette and J.~P. Reiter.
\newblock {M}ultiple {I}mputation for {M}issing {D}ata via {S}equential
  {R}egression {T}rees.
\newblock {\em American Journal of Epidemiology}, 172(9):1070--1076, 2010.

\bibitem{Cha13}
K.-W. Chang, Sundararajan S., and Keerthi S.S.
\newblock Tractable semi-supervised learning of complex structured prediction
  models.
\newblock In {\em ECML}, 2013.

\bibitem{Cra02}
K~Crammer and Y~Singer.
\newblock On the algorithmic implementation of multiclass kernel-based vector
  machines.
\newblock {\em JMLR}, 2:265--292, 2002.

\bibitem{Dhi12}
P.~S. Dhillon, S.~S. Keerthi, K.~Bellare, O.~Chapelle, and S.~Sundararajan.
\newblock Deterministic annealing for semi-supervised structured output
  learning.
\newblock In {\em AISTATS}, 2012.

\bibitem{Gha01}
Z.~Ghahramani and M.J. Beal.
\newblock Propagation algorithms for variational bayesian learning.
\newblock In {\em NIPS}, 2001.

\bibitem{Gha94}
Z.~Ghahramani and M.I. Jordan.
\newblock Supervised learning from incomplete data via an em approach.
\newblock In {\em NIPS}, 1994.

\bibitem{Joa99}
T.~Joachims.
\newblock Transductive inference for text classification using support vector
  machines.
\newblock pages 200--209, 1999.

\bibitem{Kee08}
S.S. Keerthi, S.~Sundararajan, K.-W. Chang, C.-J Hsieh, and C.-J Lin.
\newblock A sequential dual method for large scale multi-class linear svms.
\newblock In {\em SIGKDD}. ACM, 2008.

\bibitem{Kiw83}
K.~Kiwiel.
\newblock An aggregate subgradient method for nonsmooth convex minimization.
\newblock {\em Mathematical Programming}, pages 320--341, 1983.

\bibitem{Kom11}
N.~Komodakis.
\newblock Efficient training for pairwise or higher order crfs via dual
  decomposition.
\newblock In {\em Proc. CVPR}, 2011.

\bibitem{Kom11b}
N.~Komodakis, N.~Paragios, and G.~Tziritas.
\newblock Mrf energy minimization and beyond via dual decomposition.
\newblock {\em IEEE Trans. PAMI}, 33(3):531--552, 2011.

\bibitem{Kor09}
Y.~Koren, R.~Bell, and C.~Volinsky.
\newblock Matrix factorization techniques for recommender systems.
\newblock {\em IEEE Computer}, 42:30--37, 2009.

\bibitem{Li12}
F.~Li, Y.~Yu, and D.B. Rubin.
\newblock Imputing missing data by fully conditional models: Some cautionary
  examples and guidelines.
\newblock 2012.

\bibitem{Liu10}
J.~Liu, A.~Gelman, J.~Hill, and Y.-S. Su.
\newblock On the stationary distribution of iterative imputations.
\newblock {\em arXiv preprint arXiv:1012.2902}, 2010.

\bibitem{Luc13}
A.~Lucchi, Y.~Li, and P.~Fua.
\newblock Learning for structured prediction using approximate subgradient
  descent with working sets.
\newblock In {\em Proc. CVPR}, 2013.

\bibitem{Mar08}
B.M. Marlin.
\newblock {\em Missing data problems in machine learning}.
\newblock PhD thesis, University of Toronto, 2008.

\bibitem{Mes10}
O.~Meshi, D.~Sontag, T.~Jaakkola, and A.~Globerson.
\newblock Learning efficiently with approximate inference via dual losses
  (2010).
\newblock In {\em Proc. ICML}, 2010.

\bibitem{Rag01}
T.E. Raghunathan, J.M. Lepkowski, J.~van Hoewyk, and P.~Solenberger.
\newblock {A} multivariate technique for multiply imputing missing values using
  a sequence of regression models.
\newblock {\em Survey Methodology}, 2001.

\bibitem{Rat07}
N.~Ratliff, J.~A. Bagnell, and M.~Zinkevich.
\newblock (online) subgradient methods for structured prediction.
\newblock In {\em AISTATS}, 2007.

\bibitem{Su11}
Y.-S Su, A.~Gelman, J.~Hill, and M.~Yajima.
\newblock Multiple imputation with diagnostics ({mi}) in {R}: Opening windows
  into the black box.
\newblock {\em Journal of Statistical Software}, 45, 2011.

\bibitem{Sut09}
C.~Sutton and A.~McCallum.
\newblock Piecewise training for structured prediction.
\newblock {\em Machine Learning}, pages 165--194, 2009.

\bibitem{Tem11}
M.~Templ, A.~Kowarik, and P.~Filzmoser.
\newblock Iterative stepwise regression imputation using standard and robust
  methods.
\newblock {\em Computational Statistics \& Data Analysis}, 55, 2011.

\bibitem{Buu07}
S.~van Buuren.
\newblock Multiple imputation of discrete and continuous data by fully
  conditional specification.
\newblock {\em Statistical Methods in Medical Research}, 16(3):219--242, 2007.

\bibitem{Buu99}
S.~van Buuren, H~C Boshuizen, and D~L Knook.
\newblock {M}ultiple imputation of missing blood pressure covariates in
  survival analysis.
\newblock {\em Statistics in Medicine}, 18:681--694, 1999.

\bibitem{Buu11}
S.~van Buuren and K.G. Oudshoorn.
\newblock {mice}: Multivariate imputation by chained equations in r.
\newblock {\em Journal of Statistical Software}, 45(3):1--67, 2011.

\bibitem{Wai03}
M.J. Wainwright, T.S. Jaakkola, and A.S. Willsky.
\newblock Map estimation via agreement on (hyper)trees: Message-passing and
  linear programming approaches.
\newblock Technical report, EECS Dept, UC Berkeley, 2003.

\bibitem{Wes98}
J.~Weston and C.~Watkins.
\newblock Multi-class support vector machines.
\newblock {\em Technical Report CSD-TR-98-04, Dept of CS, University of
  London}, 1998.

\bibitem{Will07}
D.~Williams, X.~Liao, Y.~Xue, L.~Carin, and B.~Krishnapuram.
\newblock On classification with incomplete data.
\newblock {\em IEEE Trans. PAMI}, 29(3):427--436, 2007.

\bibitem{Win05}
J.~Winn and C.~M. Bishop.
\newblock Variational message passing.
\newblock {\em JMLR}, 2005.

\end{thebibliography}

\end{document}